# Novel sparse matrix algorithm expands the feasible size of a self-organizing map of the knowledge indexed by a database of peer-reviewed medical literature

**Andrew James Amos[1] - Kyungmi Lee[2] - Tarun Sen Gupta[1] - Bunmi S. Malau-Aduli[3]**

[1] College of Medicine & Dentistry, James Cook University, Townsville, Australia

[2] College of Science and Engineering, James Cook University, Cairns, Australia

[3] School of Medicine and Public Health, University of Newcastle, Newcastle, Australia

**A.J. Amos** is Senior Lecturer at the James Cook University College of Medicine and Dentistry; ORCID: https://orcid.org/0000-0002-9145-0212

**K. Lee** is Senior Lecturer at the James Cook University College of Science and Engineering, Cairns, Australia; ORCID: https://orcid.org/0000-0003-3304-4627

**T. Sen Gupta** is Professor of Health Professional Education, and Head, Clinical School at the James Cook University College of Medicine and Dentistry, Townsville, Australia; ORCID: https://orcid.org/0000-0001-7698-1413

**B. S. Malau-Aduli** is Professor of Medical Education and Director of the Academy for Collaborative Health Interprofessional Education and Vibrant Excellence (ACHIEVE), at University of Newcastle School of Medicine and Public Health, Newcastle, Australia; ORCID: https://orcid.org/0000-0001-6054-8498

Corresponding author: Andrew J. Amos

Andrew.Amos@jcu.edu.au

**Declarations:**

- Availability of data and material

- - All data used in the training of the Self-organizing map are available directly from Medline: https://ftp.ncbi.nlm.nih.gov/pubmed/baseline/

- Competing interests

  - All authors declare that we have no competing interests.

- Funding

  - No funding was required or provided for this research.

- Authors' contributions

  - Andrew Amos conceptualised, designed, and implemented this research

  - Bunmi Malau-Aduli, Tarun Sen Gupta, and Joanne Lee provided supervision and helped revise and edit the manuscript.

- Acknowledgements

  - No other acknowledgements to declare.

# Novel sparse matrix algorithm expands the feasible size of a self-organizing map of the knowledge indexed by a database of peer-reviewed medical literature

## Abstract

Past efforts to map the Medline database have been limited to small subsets of the available data because of the exponentially increasing memory and processing demands of existing algorithms. We designed a novel algorithm for sparse matrix multiplication that allowed us to apply a self-organizing map to the entire Medline dataset, allowing for a more complete map of existing medical knowledge. The algorithm also increases the feasibility of refining the self-organizing map to account for changes in the dataset over time.

## Keywords

## 1   Introduction

The quantity of information available to modern researchers for analysis, both structured data with defined meaning and unstructured information of uncertain meaning, is increasing exponentially.[1] However, it is impossible for any individual human being to read and understand more than a tiny fraction of the information that exists at a point in time, or to keep up with new information as it becomes available.

Fortunately, a wide variety of machine learning (ML) algorithms have been developed that can analyze and present abstract summaries of characteristics of arbitrarily large datasets relevant for particular purposes. Visualization is one of the most powerful ML techniques for condensing the information contained in large datasets into human understandable form. The most common approach is to transform high dimensional data into a flat two dimensional map which retains the properties of interest in the high dimensional space.[2]

An important example of the visualization of high dimensional data in two dimensional maps is the production of maps of scientific knowledge produced in the Science of Science field (SciSci).[3] SciSci maps have been created to summarize a wide range of relationships including research networks linking scientists, scientific units such as research labs, universities, and more abstract information such as the emergence and evolution of new scientific ideas.[4]

The most common form of SciSci visualization is citation analysis which represents abstract relationships between scientific articles published in peer-reviewed journals.[4] This type of analysis is particularly useful for identifying the evolution of scientific paradigms, comparing the relative productivity of different researchers or research units such as universities, and analyzing the impact of network dynamics such as multidisciplinarity on research productivity measured by quantity and quality.[5]

## 2. Visualization of large medical article datasets with self-organizing maps

The most comprehensive and authoritative source of information about medical knowledge and practice is contained within the articles published by peer-reviewed medical journals. While there are other stores of medical knowledge such as textbooks, manuals, and online learning materials, those types of knowledge rely upon references to peer-reviewed articles as authority for their claims.

Unlike other forms of medical knowledge, articles published in peer-reviewed journals have a systematic structure and quality-control mechanisms. In addition, they are labelled and organized to facilitate rapid identification and retrieval of articles relevant to particular information needs. All these features make the set of peer-reviewed medical articles an excellent test case for developing SciSci visualization techniques.

Skupin has provided the most advanced visualizations of the peer-reviewed medical literature. His maps combine the powerful pattern recognition of self-organizing maps (SoMs) with the sophisticated visual cues of map-making to produce high density visual abstractions of the domains of medical knowledge (Figure 1).[6]

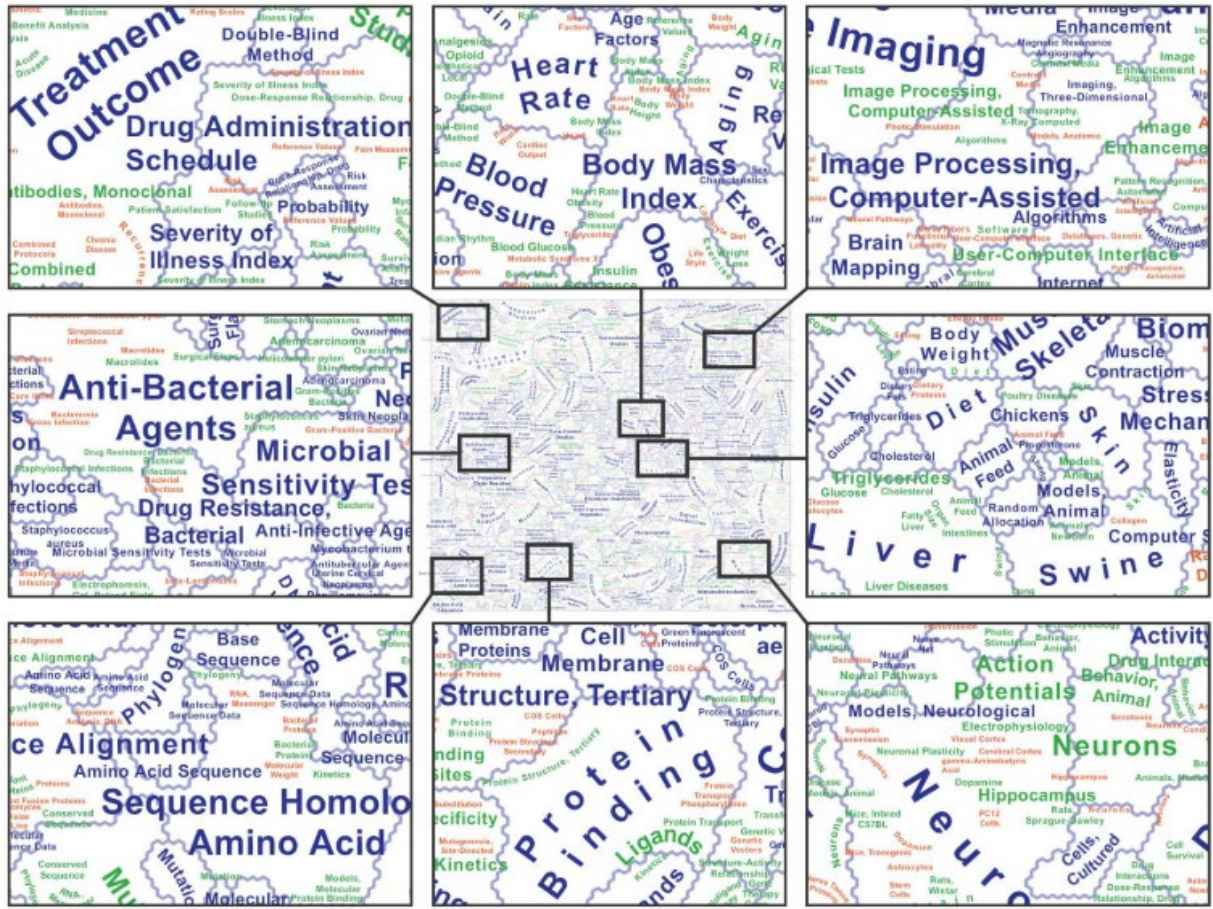

**Figure 1: Skupin's map of the medical literature showing regions of interest is licensed under CC-BY (https://creativecommons.org/licenses/by/4.0/)**[6]

Due to hardware and software limitations, Skupin's SoM was trained on 2.1 million of the more than 20 million indexed peer-reviewed articles available at the time. In addition, the size of the SoM itself was limited to a two dimensional grid of neurons of size 275 x 275 for a total of 75,625 neurons. Skupin conceded that it would have been preferable to increase the dimensions of the grid to have a smaller article to neuron ratio and improve the resolution of the map, but that this was not computationally feasible.[6]

Skupin et al.[6] acknowledged that the size of their map limited its ability to reproduce detailed structures of the high-dimensional knowledge domain on the two-dimensional surface of the SoM. In addition to the limitation of being forced to train the SoM on a sample of about 10% of articles rather than the complete set, they were only able to consider a small subset of the medical subject headings (MeSH – described below) used to annotate each article. As a result, the Skupin SoM is unable to represent any information that was contained in the 90% of articles and MeSH not used for training.

Another limitation of a small SoM and partial training set is that it constrains the complexity of the relationships that can be represented. Figure 1 illustrates this point.[6] Each of the 8 rectangles surrounding the central map represents a cluster of related knowledge structures. For example, the bottom right rectangle represents the high level specialty Neurology, superimposed on lower level anatomical structures like neurons and hippocampi, as well as neural processes like action potentials.

The exclusion of most of the MeSH from training means that their map cannot contain most of the categories that annotate Medline-indexed articles. For example, the Skupin map might contain the high level MeSH structure Hippocampus, but not the lower-level MeSH structure CA1 Region, Hippocampus; or it might contain the high level MeSH-labelled process Action Potentials, but not the related process of Synaptic Potentials. As the MeSH controlled vocabulary has a hierarchical structure with high level concepts branching into multiple lower level concepts, this necessarily means that the Skupin SoM is limited to representing relationships between relatively high level MeSH.

A final limitation imposed by computational feasibility is the inability to extend the SoM to consider more than two dimensions. While it is technically possible to add a third spatial dimension to a SoM, it requires special equipment to make use of the results.[7] However, the ability to consider how a two dimensional spatial SoM changes over time promises to reveal the dynamic processes involved in the emergence and evolution of knowledge structures.[8] Any technique that increases the computational efficiency of self-organising map training also improves the possibility of extending the regular model to reveal new processes such as temporal relationships.

## 2.1 Visualization with self-organizing maps

The self-organizing map pioneered by Kohonen has been used to visualize subsets of the Medline database which indexes and annotates more than 30 million articles from high-quality peer reviewed medical journals. Kohonen's iterative algorithm has many attractive properties for generating visualizations of large sets of scientific knowledge. It can effectively map very high dimensional information spaces to two dimensions while retaining many of the topological relationships of the higher dimensional space. In addition, it can detect and help visualize previously unknown relationships.[2]

Self-organizing maps are single-layer neural networks which transform high-dimensional vectors into low-dimensional vectors (usually 2D) while retaining many of the topological properties of the higher dimensional space (Figure 2).[2,9] The most obvious example of the retention of topological properties by SoMs is that units of information that are close together in high-dimensional space will also be close together in the 2D space represented by the SoM.

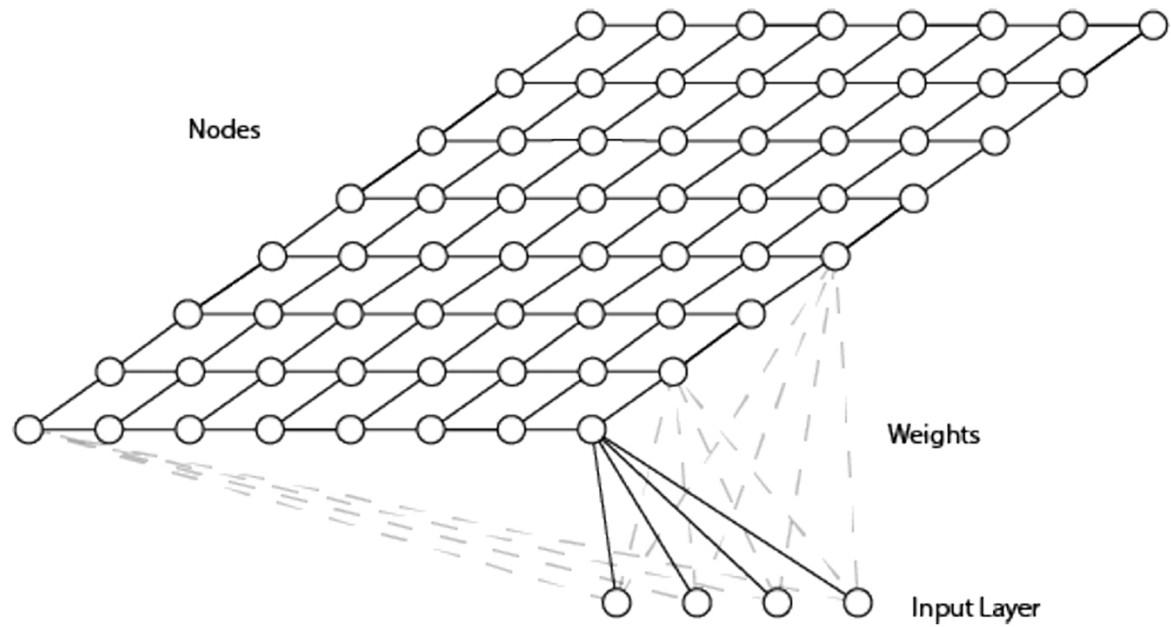

**Figure 2: Self-organizing map with a 4-dimensional input layer and a 2-dimensional output layer of 64 nodes**

This retention of topological properties by SoMs is extremely useful for visualization because human beings have visual systems that are highly developed for discriminating spatial relationships in 2D space but have almost no ability to perceive similar relationships in high-dimensional space. As a result, SoMs can make complex relationships from high dimensional spaces intelligible to human vision.

Kohonen[2] described two algorithms for training SoMs to transform high-dimensional input vectors to 2D neuronal maps. Each neuron in the SoM is represented by a vector of weights of the same dimensionality as the input vectors. Before training the weight vectors of all neurons are filled with random values. Training occurs in distinct epochs during which every input is separately presented to the entire network of neurons and the distance between each input vector and each neuron is calculated by a standard distance function, often the squared Euclidean distance. Each input is linked with the neuron that is closest to it, and training is implemented by iterative changes that move neuronal weight vectors towards the inputs that they are closest to.

In the standard SoM algorithm training changes are made to weights after every individual input is presented, while in the batch SoM algorithm changes are accumulated over the entire epoch and applied at the end of the epoch. The batch algorithm is computationally more efficient than the standard algorithm, and is thought to converge on the same set of training outcomes, although this has not been formally proved.[2]

## 2.2 Improving computational efficiency for sparse input vectors

Other techniques are used to improve the computational efficiency of SoM algorithms for specific types of input. A technique that is useful for mapping the medical literature is a form of the batch algorithm that is optimised for sparse inputs. An input vector can be considered sparse when most of the input elements are empty (or where most of the elements have a single constant value such as 0 or 1 that can effectively be treated as empty).[10,11]

Sparse inputs present a challenge for GPU-based computation because they complicate the data-partitioning that is required to distribute tasks across the large number of processing units. Melka & Mariage[10,11] described a modified version of the batch SoM algorithm optimised for sparse inputs and suitable for GPU processors (Figure 3). Essentially their algorithm depends on the fact that only non-zero input elements affect the weights of each neuron in the SoM which allows them to precompute the values of the squared norms of the distance equation once for the input vectors, and once per epoch for the weights of each neuron, substantially reducing the computational load.

| | **Variables/Data structures** |
|---|---|
| | **Input:** $x$ a set of $N$ sparse vectors of $D$ components |
| | **Data:** $\omega$ initialized codebook of M dense vectors |
| | **Data:** $\chi$ an array of $N$ reals, satisfying $\sum_j x_{ij}^2$ |
| | **Data:** *dst* array of N reals to store best distances |
| | **Data:** *bmu* array of N integers to store best match units |
| | **Data:** *nmu* array of D reals to accumulate numerator values |
| **Algorithm** | |
| 1 | **for** $i \leftarrow$ 1 **to** $N$ **do** $\chi_i \leftarrow \sum_j x_{ij}^2$; init $\chi$ |
| 2 | **For** $e \leftarrow$ 1 **to** $e_{max}$ **do** train one epoch |
| 3 |     interpolate $\sigma$; |
| 4 |     **for** $i \leftarrow$ 1 **to** $N$ **do** $dst_i \leftarrow \infty$ ; initialize *dst* |
| 5 |     **for** $k \leftarrow$ **to** $M$ **do** find all bmus |
| 6 |         $\omega \leftarrow \sum_j x_{kj}^2$ |
| 7 |         **forall** $i \in 1,\ldots,N$ **do** |
| 8 |             $d \leftarrow \omega + \chi_i - 2(x_i \cdot \omega)$; |
| 9 |             **if** $d < dst_i$ **then** store best match unit |

| 10 | $dst_i \leftarrow d$; |
| 11 | $bmu_i \leftarrow k$; |
| 12 | **forall** $k \in 1,...,M$ **do** |
| 13 | $den \leftarrow 0$ ; init denominator |
| 14 | **for** $j \leftarrow 1$ **to** $D$ **do** $num_j \leftarrow 0$; init numerator |
| 15 | **for** $i \leftarrow 1$ **to** $N$ **do** accumulate $num$ and $den$ |
| 16 | $c \leftarrow bmu_i$ ; |
| 17 | $h \leftarrow \exp(\frac{\|r_k - r_e\|^2}{2\sigma^2}$ |
| 18 | $den \leftarrow den + h$ ; |
| 19 | **for** $j \leftarrow 1$ **to** $D$ **do** |
| 20 | $num_j \leftarrow num_j + hx_{ij}$ |
| 21 | **for** $j \leftarrow 1$ **to** $D$ **do** update $\omega_k$ |
| 22 | $\omega_{kj} \leftarrow \frac{num_j}{den}$ |

**Figure 3: Sparse batch SoM algorithm**[10,11]

## 3.   Optimising the SoM sparse algorithm for nominal inputs

### *3.1   The Medline database*

The SoM algorithm described by Melka & Mariage[10,11] is designed for sparse inputs comprising ratio variables (ie real numbers with arbitrary limits). Some data sets involve sparse inputs with nominal variables which allow for further optimization of the batch algorithm that increase computational efficiency and decrease the memory requirements for storing and distributing the input vectors.

The Medline database used by Skupin to create SoMs of the knowledge contained within the peer-reviewed medical literature is an example of a database with nominal input vectors.[6] The Medline database indexes a detailed set of information for each article published in a selected group of high-quality peer-review medical journals. It includes articles all the way back to the 19th century, but coverage is limited prior to the 1970s.[12]

For each indexed article Medline records the title, abstract, year of publication, and authors, alongside many other variables.[13] Most relevant to the SoMs produced by Skupin, Medline annotates every published article with a set of descriptors called Medical Subject Headings (MeSH) from a controlled vocabulary maintained by the National Library of Medicine. MeSH describe the content, methods, and other characteristics of each article (such as research design, patient population, and disease treated).

For example, an article describing a treatment trial where patients with schizophrenia were treated with the antipsychotic medication risperidone and compared with controls treated with placebo would be annotated with MeSH including the diagnosis of Schizophrenia, the experimental design Randomised Control Trial, and the treatment group Antipsychotic – Risperidone.

Skupin's SoM used the MeSH annotating each article in the Medline database as an input vector, where the nominal MeSH variables were represented as a vector of binary elements.[6] Skupin selected the most frequently used 10% of the 23,000 MeSH in the controlled vocabulary for the inputs to his SoM. Each article in the Medline database was then characterised by a vector with 2,300 elements that were coded as either present or absent (numerically as 0 or 1). As articles in the Medline database are annotated with an average of 10 MeSH, the input vectors were highly sparse, with around 2,290 elements indicating a MeSH was absent and around 10 elements indicating a MeSH was present.

Compared with the real variable inputs to Melka & Mariage's batched sparse SoM,[10,11] the binary inputs of the Medline dataset represent an additional opportunity for optimisation. While various structures are used for compactly storing sparse matrices and algebraically combining them with other sparse and dense matrices, they generally require at least two vectors – one to describe the position of a non-zero element, and a second to describe its value. The binary input vectors representing Medline articles can dispense with the second vector.

In addition, because all elements in the sparse Medline input vectors are either zero or one, the dot-product used to calculate distances in the SoM algorithm can be replaced by simple addition. Combining these two modifications, Amos et al.[14] described an algorithm (Figure 4; the Batched sparse binary SoM - MedSoM) that increased the computational efficiency and data compression sufficiently to create a SoM trained on the entire set of more than 33 million articles indexed in Medline as of 01/01/2021; and included all 29,917 of the MeSH in the NLM's controlled vocabulary.

| **Variables/data structures** | |
|---|---|
| | **Input**: $x$: $N$ sparse vectors of $D$ elements representing Medline articles |
| | **Data:** $\omega$: initialized codebook of $M$ dense vectors of length $D$; represents weights between inputs and each SoM node |
| | **Data:** $\chi$: array of $N$ reals; as $x$ is sparse this matrix is the number of non-zero elements per row (i.e., the number of MeSH annotating each article) |
| | **Data:** $dst$: array of $N$ reals storing distance to best matching unit for each article |
| | **Data:** $bmu_1$, $bmu_2$ – 2 vectors of $N$ integers to store best matching and second-best matching unit/node for each article |
| | **Data:** $num$ – array of $D$ reals to accumulate numerator values |
| **Algorithm** | |
| **1** | **Randomize** codebook weights $\omega$ between 0.0f and 1.0f |
| **2** | **for** each epoch $e \leftarrow 1$ to $K$ **do** |

| | | |
|---|---|---|
| 3 | **compute** $\sigma$; | // radius for current epoch |
| 4 | **standardize** codebook weights; | sqrt of sum of weights per node |
| 5 | **calculate** $bmu_1$, $bmu_2$ for each article $i \leftarrow 1$ to $N$ | |
| 6 |     **for** each node $m \leftarrow 1$ to $M$ | |
| 7 |         **sum** $m \cdot i$ | // dot-product replaced by sum as all $i = 1$ |
| 8 |         **calculate** $dst = D_a - 2 \cdot (m \cdot i)$ | |
| 9 | **calculate** new weights for each node $\leftarrow 1$ to $M$ **do** | |
| 10 |     **for** each article $i \leftarrow 1$ to $N$ | |
| 11 |         **calculate** $h(e) = \exp\left(-\frac{\|r_k - r_c\|^2}{2\sigma(e)^2}\right)$ | // Neighborhood function |
| 12 |         **reduce** denominator = accumulate $h_{ck}$ per node | |
| 13 |         **calculate** numerators = accumulate $h_{ck}$ per node/article | |
| 14 |         **calculate** new weights = numerators / denominator | |
| 15 | **for** $i \leftarrow 1$ to N | |
| 16 |     **calculate** adjacency = ($bmu_1 - bmu_2 =< 1$) | // Manhattan distance |
| 17 |     **calculate** topographic error = % adjacent | |

D – total number of MeSH categories (=29,917); $D_a$ – number of MeSH categories annotating an individual article; M – number of nodes (350 x 350 = 122,500); N – number of articles (=33,375,866); $h(e)$ – width of the neighborhood (which changes over training epochs according to the formula $175/(1.7)^{epoch}$ )

**Figure 4: Sparse batch SoM algorithm for binary/nominal input vectors (MedSoM)**

Starting with the 275 x 275 = 75,625 neuron SoM used by Skupin to model the subset of 2.1 million articles with 2,300 MeSH, Amos et al. experimented with different sizes of SoM in increments of 25 x 25 up to 400 x 400 = 160,000 neurons, and settled on a 350 x 350 = 122,500 neuron SoM. Maps larger than this did not achieve any greater reduction in topographic error, which measures the % of articles where the best matching unit and second-best matching unit/node are not adjacent.[15]

### 3.2  Iterative improvement of the SoM algorithm

The LibSVM and MedSoM algorithms seek to improve the efficiency of the SoM algorithm by using the properties of sparse matrix multiplication. In addition, by reducing the space required to represent the training articles in GPU memory, the MedSoM algorithm makes it possible to leverage GPU specific optimisations of matrix multiplication.

The current implementation of all the algorithms used in this research use general memory accessible to all GPU threads rather than the cache memory accessible only within thread blocks. This substantially slows processing speed because it requires atomic calls to article information in general memory for each MeSH representing each article.

In general, the speed of implementation of GPU algorithms in programming kernels is considered to be either compute bound, where the time taken to complete each instruction set depends mainly on the time taken to complete the computational instructions; or memory bound, where the time taken to complete each instruction set depends mainly on the time taken to copy the data required to complete the instructions from other parts of memory to the thread block where it is required.[16]

Reducing the size of memory required to represent articles by moving from dense to sparse representation of articles makes it feasible to optimise processing by copying the representation of articles into each thread block, eliminating the need for atomic operations that pause computation while retrieving individual items from general memory. Reducing the memory requirement further by moving from the LibSVM to the MedSoM algorithm increases the number of articles that can be copied into the local cache of each thread block. Both reduce the possibility/extent of being memory bound and compute bound for any given set of Medline data.

### 3.2   Capitalizing on improved performance with model extensions

In addition to making larger and more complex SoMs possible, increasing the speed and decreasing the memory requirements of the SoM make it more feasible to create extensions that expand the capacities of the basic model. One of the main limitations of the existing MedSoM is that it was trained with articles across more than a century of research. While it is likely there were some features of medical practice in 1900 that remain true today, it is certain that there are many features that have changed radically. The MedSoM currently has no way of differentiating between such features.

One extension of the SoM that would allow MedSoM to detect changes in the peer-reviewed medical literature over time is Denny et al.'s relative density SoM (RedSoM).[8] By training a SoM on each year of their dataset and then analyzing differences in the map between years, Denny et al. were able to visually identify "emerging clusters, disappearing clusters, split clusters, merged clusters, enlarging clusters, contracting clusters, the shifting of cluster centroids, and changes in cluster density" (p281).[8] For example, by this technique it would be possible to identify the emergence of clusters of new research activity, and the disappearance of clusters of old research activity.

The ultimate goal of Amos et al.'s[14] MedSoM was to provide a tool capable of providing an empirical basis for curriculum development, in order to address the biases inherent in the current reliance on expert judgement. While MedSoM's knowledge map of the entire set of information contained in more than a century of peer reviewed medical literature was the first step towards this goal, the ability to differentiate between emerging and obsolete information is clearly vital for informing a curriculum.

## 4.   Performance evaluation

The optimised algorithm described by Amos et al.[14] is important for two reasons. By decreasing the size of the sparse input vector it increases the size of the dataset that can be physically held in memory and used to train the SoM describing the published medical knowledge. By simplifying the calculation of the distance between each article and each neuron in the SoM it increases the size of the SoM that can be trained in realistic time by current technology.

We set out to demonstrate both these factors by comparing the performance of the Amos et al.[14] MedSoM algorithm to Melka & Mariage's[10,11] BSoM algorithm by calculating the maximum size of the input/weight vectors, SoM, and article input set possible with each. We then compared the speed performance of each algorithm on an epoch of training using Amos et al. as the baseline. We focused on the matrix multiplication required to calculate the best matching unit for each article, which was the most computation- and time-intensive part of the algorithm.

## 4.1 Technical specifications

The analyses were performed on a Windows 11 PC with an Intel(R) Core(TM) i7-14700F 2.10 GHz CPU, 32GB RAM, and an NVIDIA GeForce RTX 4090 with 24GB RAM GPU.

## 4.2 Memory

To analyse the influence of the sparse Melka and MedSoM algorithms on the feasible size of the SoM compared with a dense SoM algorithm we created a series of self-organising maps and sets of articles represented either by dense matrices of 4-byte floats with most elements set to 0.0 and annotated MeSH set to 1.0 (Dense-SoM); or by sparse matrices in either the Lib-SVM format which includes one 4-byte integer that identified each present MeSH, and a 4-byte float that represented the value of the MeSH; or the MedSoM format that included one 4-byte integer that identified each present MeSH. Each article was annotated with a random number of MeSH continuously distributed between 5 and 15.

Within this framework we compared the tradeoffs and limits between the number of articles, number of MeSH, and SoM dimensions, across the dense and sparse algorithms. The results are presented in Figure 5.

When interpreting Figure 5 it is important to note that 5(a) is different from 5(b,c) in two ways. In order to illustrate the differences between the memory usage of the sparse algorithms, 5(a) represents the full range of 30 million articles. As can be seen, this flattens the memory usage of the Dense algorithm against the upper left limits of the graph, making it impossible to understand the tradeoffs between MeSH number and article number; and it makes it difficult to see the memory usage of the Sparse algorithms for Articles, barely perceptible as a thin line at the bottom of the left limits of the graph.

In order to see the tradeoffs between MeSH and articles for the Dense algorithm, 5(b) limits the range of articles between 0 and 1,200,000. As 3-dimensional graphs are easier to read when the figures are concave, the axes representing MeSH and Articles are switched. In order to see the differences in memory usage for the storage of articles by the Sparse algorithms, 5(c) limits the range of articles between 0 and 1,200,000, limits the range of memory usage from 0 to 110mb, and switches the MeSH and Article axes.

With those modifications in mind, 5(a) shows that both sparse algorithms enormously increase the capacity of self-organizing maps to represent both an increased number of Articles and an increased number of MeSH. For a SoM of dimensions 350x350, on the research equipment with 24GB of memory available, the dense algorithm was able to represent 1,000,000 articles annotated with 5,000 MeSH; or to represent 70,000 articles with 30,000 MeSH.

Figure 5(b) shows the tradeoffs for the dense algorithm between number of MeSH and number of Articles, with the memory used to store the SoM increasing and memory used to store Articles decreasing as the dimensions of the SoM increased from 250x250 to 350x350.

Figure 5(a) also shows that the MedSoM algorithm significantly decreases the memory used to store Articles compared with the LibSVM algorithm. For a SoM of 350x350 nodes, representing 30,000 MeSH, and trained on 30,000,000 articles, the MedSoM algorithm decreased the memory used from 3372mb to 1378mb relative to the LibSVM algorithm, with proportional decreases across the range of MeSH, articles, and SoM dimensions, a 59% reduction. Both sparse algorithms used the same memory to store the SoM across all dimensions.

Figure 5(c) shows that this advantage for MedSoM extended across the entire range of SoM dimensions, number of MeSH, and number of articles, with a 50% reduction from 12mb to 6mb to store articles with a SoM of 250x250 nodes, 30,000 MeSH, and 120,000 articles.

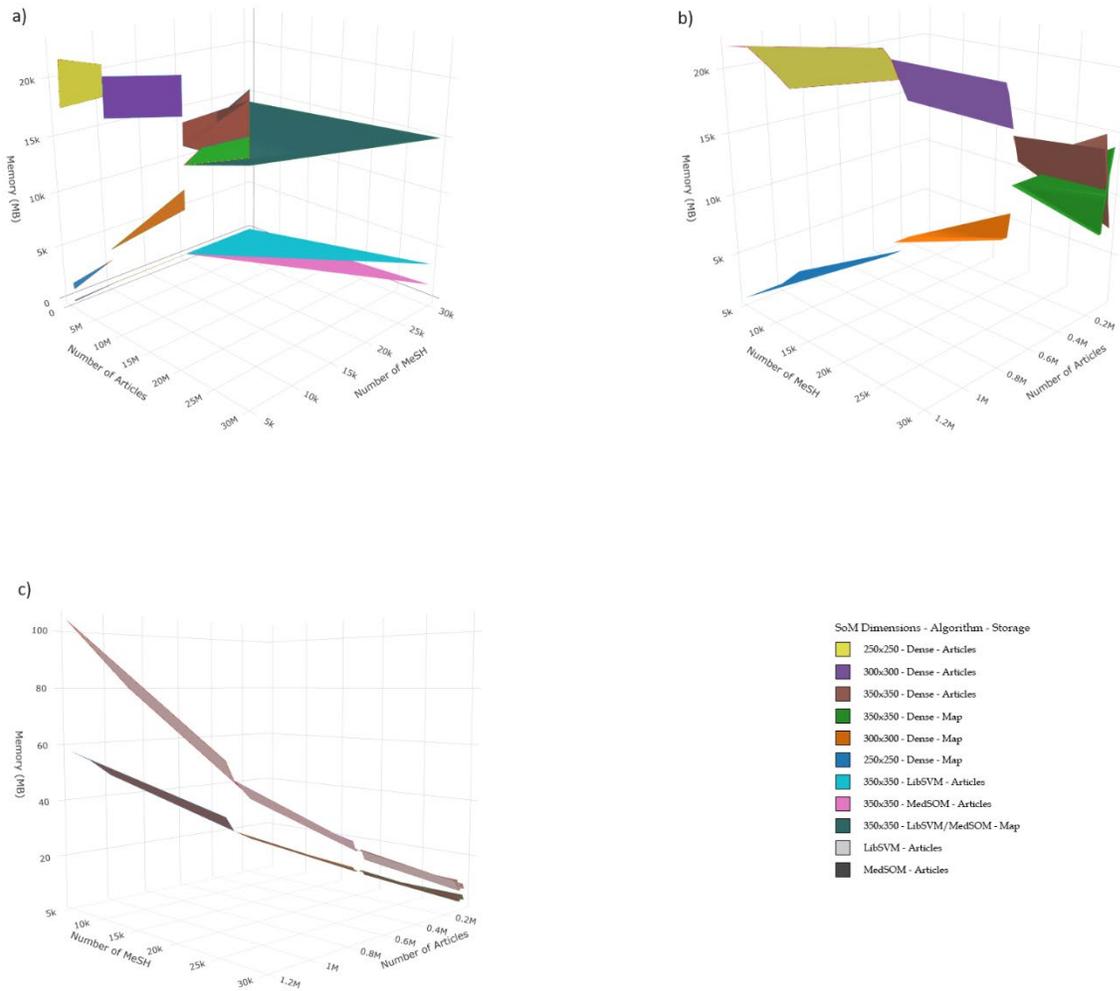

Figure 5 – Memory storage required for sparse and dense algorithms across conditions: a) Compares MB required to store articles and SoM for dense and sparse algorithms; b) Shows MB required to store articles and SoM for dense algorithm alone; c) Compares MB required to store articles for sparse algorithms

## 4.3 Processing time

To analyse the influence of the sparse Melka and MedSoM algorithms on processing time we compared the time required to complete a single cycle of the matrix multiplication used to calculate the best-matching unit for each article (($x_i \cdot \omega$) on line 8 of Figure 3 and *($m \cdot i$)* on line 7 of Figure 4). The results are reported in Figures 6 & 7.

Figure 6 shows the massive advantage of the sparse algorithms over the dense algorithms in terms of processing speed. For a SoM of 350x350 nodes, memory constraints prevent it from handling more than 450,000 articles with 10,000 MeSH, or 70,000 articles with 30,000 MeSH. While the dense algorithm took 178 minutes to process one cycle of the best matching unit dot-product with these parameters, MedSoM took less than 1 minute. The same tradeoffs between MeSH and article numbers seen for memory are replicated for processing speed, shown by the reduction in processing time per cycle to 90 minute gained by changing the article number to 70,000 and the MeSH to 30,000.

At the lower end, while the dense algorithm took 9 minutes to process a 350x350 node SoM with 50,000 articles and 5,000 MeSH, MedSoM took less than 1 second (554 times faster). Note that differences between the MedSoM and LibSVM algorithms are not visible at the scale shown by Figure 6, so MedSoM is used to represent both.

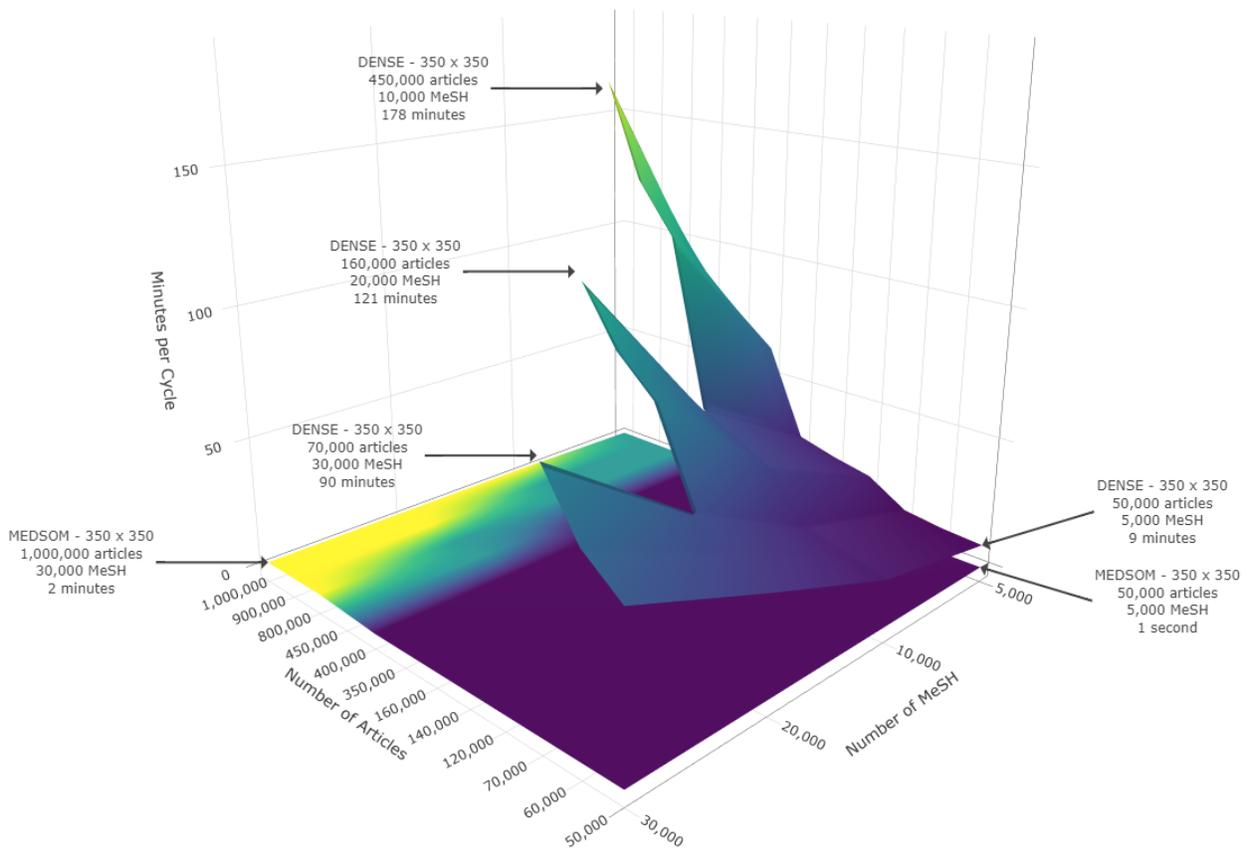

Figure 6: Processing time in minutes per cycle for the dense and MedSoM algorithms

Figure 7 shows that the MedSoM algorithm maintained a processing speed advantage across all SoM dimensions, MeSH numbers, and article numbers. MedSoM was 5% faster (81 versus 85 minutes) than LibSVM at 350x350, with 32,000 MeSH and 32 million articles, and 5% faster at 350x350 with 32,000 MeSH and 8 million articles. Interestingly, MedSoM's advantage was larger with smaller datasets, taking 1927 milliseconds compared to 2,567 milliseconds at 50x50 with 1,000 MeSH and 8 million articles, a 25% reduction.

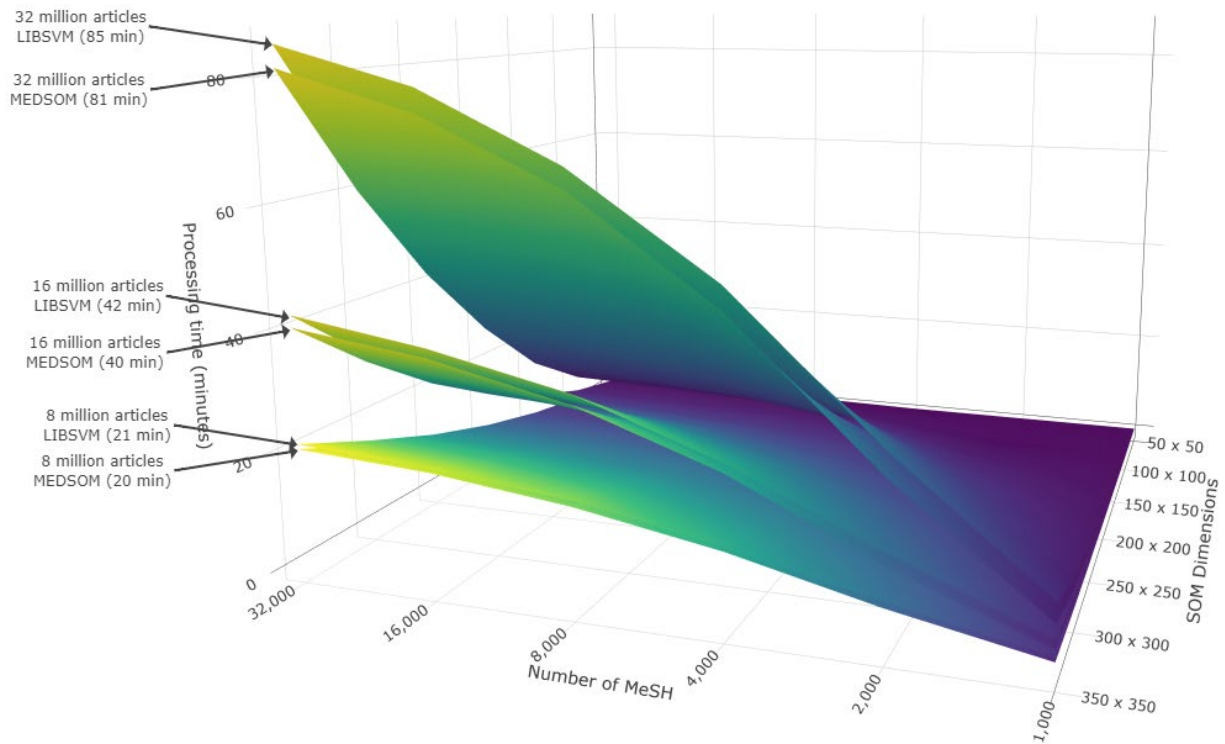

Figure 7: Processing time in minutes per cycle for the LibSVM and MedSoM algorithms

## 5.    Discussion

The power of the SoM is to condense and present in human-readable form representations of the large sets of information indexed by the Medline database. Not only does the technical innovation of the MedSoM make it possible to represent the entire database of articles rather than a subset, it makes it possible to represent all the different MeSH that describe different articles. As it is not possible to predict beforehand which patterns may emerge from what subsets of articles and MeSH, a SoM that represents the entire set of knowledge can be assumed to have the best chance of identifying the full set of meaningful patterns contained in the high-dimensional space of the Medline database.

Over and above the order-of-magnitude improvements in memory use and processing time moving from the dense algorithm to the sparse algorithms, MedSoM reduced the memory required to store the training articles by ~50%, and reduced processing time by ~5%. These improvements are important in their own right, as they increase the capacity of the SoM to accommodate the exponentially increasing number of articles indexed by the Medline database. In addition, the 50% reduction in size of the memory used to store articles makes possible further optimizations at the level of the kernels used to implement the SoM algorithms on GPUs.

As discussed in section 3.2, in their current implementation both sparse algorithms are primarily memory bound, particularly because of the use of atomic operations to retrieve article information from general memory. By halving the memory required to represent each article, MedSoM makes it possible to copy twice as many articles to each thread block, eliminating the use of atomic operations, which has the potential to substantially improve processing speed.

In addition, by increasing the capacity of the SoM to incorporate the entire set of information including articles from across the 150-year history of Medline, it becomes possible to do more sophisticated analyses. Most importantly, it becomes feasible to examine changes in the organisation of published medical knowledge over time using techniques such as Denny et al.'s RedSoM.[8]

## 6. Conclusions

Self-organizing maps are a valuable tool for identifying and visualizing previously unknown but extremely complex relationships in large databases of highly dimensional data. The current research reports a sparse algorithm that allowed for the training of a 350x350 node SoM on the entire dataset of more than 30 million peer reviewed medical articles indexed by the Medline database with the complete set of 29,917 MeSH, compared with a dense algorithm that was limited to a 275x275 node SoM and trained on 2.1 million articles with 2,300 MeSH.

This incremental step in the capacity of the SoM technique to provide an intelligible map of the entire set of medical knowledge contained in the published literature also points towards the next step, which will build upon the MedSoM in order to provide an account of the changes in information over time.

## Data availability Statement

Data sets generated during the current study are available from the corresponding author on reasonable request. All of the input training data is publicly available from Medline by direct download: https://ftp.ncbi.nlm.nih.gov/pubmed/baseline/ (other options including FTP are available).